\documentclass[letterpaper, 10 pt, conference]{ieeeconf}  

\IEEEoverridecommandlockouts                              

\usepackage{amsmath, amssymb}   
\usepackage{amsthm}                 
\newtheorem{definition}{Definition} 

\usepackage{algorithm}          
\usepackage{algpseudocode}      
\usepackage{threeparttable}     
\usepackage{multirow}           
\usepackage{booktabs}           
\usepackage{hyperref}           
\usepackage{graphicx}

\title{\LARGE \bf
Differentiable Predictive Control for Robotics: A Data-Driven Predictive Safety Filter Approach
}

\author{John Viljoen$^{1}$, Wenceslao Shaw Cortez$^{2}$, J\'an Drgo\v na$^{2}$, Sebastian East$^{3}$, Masayoshi Tomizuka$^{4}$, Draguna Vrabie$^{2}$
\thanks{*This research was partially supported by the Laboratory Directed Research and Development (LDRD) investments at Pacific Northwest National Laboratory (PNNL). PNNL is a multi-program national laboratory operated for the U.S. Department of Energy (DOE) by Battelle Memorial Institute under Contract No. DE-AC05-76RL0-1830.
}
\thanks{$^1$John Viljoen is a PhD student in the Department of Mechanical Engineering, University of California Berkeley, 2521 Hearst Ave, Berkeley, CA, USA
        {\tt\small john\_viljoen@berkeley.edu}}
\thanks{$^{2}$W. Shaw Cortez, J. Drgona, and D. Vrabie are with Pacific Northwest National Laboratory, 902 Battelle Blvd, Richland, WA, USA
        {\tt\small {\{w.shawcortez, jan.drgona, draguna.vrabie\}}@pnnl.gov}}
\thanks{$^3$At the time this work was conducted, Sebastian East was with the Department of Aerospace Engineering, University of Bristol, Queen's Building, Bristol, BS8 1TR 
        {\tt\small research@sebastianeast.com}}
\thanks{$^{4}$Masayoshi Tomizuka is a faculty member of the Department of Mechanical Engineering, University of California Berkeley, 2521 Hearst Ave, Berkeley, CA, USA
        {\tt\small tomizuka@berkeley.edu}}
}

\begin{document}

\maketitle

\begin{abstract}
Model Predictive Control (MPC) is effective at generating safe control strategies in constrained scenarios, at the cost of computational complexity. This is especially the case in robots that require high sampling rates and have limited computing resources. Differentiable Predictive Control (DPC) trains offline a neural network approximation of the parametric MPC problem leading to computationally efficient online control laws at the cost of losing safety guarantees. DPC requires a differentiable model, and performs poorly when poorly conditioned. In this paper we propose a system decomposition technique based on relative degree to overcome this. We also develop a novel safe set generation technique based on the DPC training dataset and a novel event-triggered predictive safety filter which promotes convergence towards the safe set. Our empirical results on a quadcopter demonstrate that the DPC control laws have comparable performance to the state-of-the-art MPC whilst having up to three orders of magnitude reduction in computation time and satisfy safety requirements in a scenario that DPC was not trained on.
\end{abstract}

\section{INTRODUCTION}

Model Predictive Control (MPC) has been applied successfully in a wide range of robotics applications \cite{data_driven_quads, mpc_legged_bots, mpc_quadrupeds}. It works by solving a constrained optimization problem online at discrete time intervals.  MPC utilizes a model of the system it is operating within to predict its future behavior over a finite horizon. At each time step, the MPC will optimize a series of control actions from the current time to the end of the horizon to reduce a cost function dependent on the predicted states and control actions \cite{mpc_textbook}.

Reinforcement learning (RL) offers significant runtime advantages over MPC because no online optimization has to be done, however, it is usually data-expensive. Differentiable control techniques offer a way to lower the data requirements of RL significantly and have recently become a popular area of research \cite{differentiable_mpc, deluca, diffstack, doc, barrier_net, pontryagin}. Differentiable Predictive Control (DPC) is a learning-based controller that uses the MPC cost function and a differentiable system model to compute policy gradients directly. The constraints in DPC are enforced using penalty methods \cite{drgona2020learning, drgovna2021deep, drgovna2022differentiable}. The policy is trained via gradient descent through the MPC cost function, the constraint penalties, and, most importantly, the unrolled dynamics of the system. This is equivalent to differentiating the Lagrangian reformulation of the optimal control problem across rollouts of the system. This method has a strong learning signal and small data requirements due to the existence of gradients from the policy to the unrolled dynamics. DPC, therefore, can be thought of as moving the optimization stage of the MPC from online to offline, which is particularly attractive for mobile robotics which are compute constrained.

Dynamics common in robotics applications often do not have a ``well-defined" Vector Relative Degree (VRD) (\cite{vrd_book}, Definition 1). This creates difficulties for gradient-based nonlinear control techniques \cite{relative_degree_feedback_lin, bad_zero_dynamics}, including learning-based controllers. VRD is effectively the number of steps a discrete-time dynamical system must take before we ``see" its input from its output. If the VRD of a system is discontinuous across the state-space, then we say it is ``poorly-defined". A poorly-defined VRD is problematic for model and gradient-based optimization methods such as DPC because if there is no gradient through the dynamics then there is no learning signal for it to operate on. In nonlinear control there are countermeasures to this, such as state augmentation in feedback linearization \cite{nonlinear_control_systems}, and system decomposition in backstepping \cite{nonlinear_and_adaptive_control_design}. In this paper, we develop a novel technique that systematically decomposes the dynamics of a system to circumvent the problem of poorly defined VRD. 

Learning-based controllers typically lack strict constraint enforcement when represented as a black-box Neural Network (NN). Recent work has aimed to tackle this by developing probabilistic guarantees of safety \cite{Hertneck8371312, Qibo10329471, Paulson9034170} and using Predictive Safety Filters (PSF) \cite{Tearle9484747, shawcortez2022differentiable}. PSFs are inspired by MPC and use predictions of future system states and predefined safety rules to decide whether to modify proposed actions by the control system, thereby ensuring the system operates within (or converges to) safe bounds \cite{predictive_safety_filter}. Defining these safety rules is challenging. Sampling-based approaches have proven effective \cite{sampling_clbf}, and convex hulls have proven effective at representing safe sets in convex environments \cite{convex_hull_safe_set}. This paper develops a novel way of defining these safety rules by forming a data-driven safe set in non-convex environments from the DPC training dataset. 

This paper also develops a novel PSF formulation to keep computational cost orders of magnitude below MPC. An event-trigger similar to \cite{shawcortez2022differentiable} is used that operates on the safe set itself. It predicts the nominal control using the Jacobian of the learned controller and uses approximate tangent planes of the safe set to keep the computation tractable. This system approximately converges the DPC-controlled system to the region of the state space upon which it was trained, preventing distribution shift problems \cite{distribution_shift}. The contributions of this paper are:

\begin{enumerate}
    \item application of \textbf{dynamics decomposition} to quadcopter dynamics to \textbf{ensure DPC learning signal is well-defined}
    \item novel use of DPC dataset for \textbf{safe set generation}
    \item novel PSF to \textbf{ensure} \textbf{safety} and \textbf{remove} \textbf{distribution} \textbf{shift} while \textbf{remaining} \textbf{computationally} \textbf{cheap}
    \item \textbf{open source} code base \url{https://github.com/pnnl/dpc_for_robotics}
\end{enumerate}

\section{Preliminaries}

Unless otherwise specified variables in \pmb{bold} ($\pmb{x}$) refer to vectors, and non-bold ($x$) refer to scalars, and non-bold with subscript ($x_i$) refers to the $i$th element of a vector ($\pmb{x}$) or the $i$th output of a function. For time-varying variables in discrete time, their value at time $k$ is represented as $(.)[k]$.

Consider the Multi-Input Multi-Output (MIMO) nonlinear discrete-time control-affine system described by:
\begin{equation}
    \begin{aligned}
        \pmb{x}[k+1] &= f(\pmb{x}[k]) + g(\pmb{x}[k]) \pmb{u}[k], \\
        \pmb{y}[k] &= h(\pmb{x}[k]),
    \end{aligned}
    \label{eqn:mimo_example}
\end{equation}

where \(\pmb{x}[k] \in \mathbb{R}^n\) represents the state of the system at time step \(k \in \mathcal{K}\), \(\pmb{u}[k] \in \mathbb{R}^m\) is the control input, \(\pmb{y}[k] \in \mathbb{R}^l\) is the output, and \(f:\mathbb{R}^n \rightarrow \mathbb{R}^n\), \(g:\mathbb{R}^n \rightarrow \mathbb{R}^{n \times m}\), and \(h: \mathbb{R}^n \rightarrow \mathbb{R}^l\) are nonlinear functions. The set \(\mathcal{K} \subseteq \mathbb{Z}\) denotes the discrete-time indices for which the system is defined.

\begin{definition}[Vector Relative Degree]
    The Vector Relative Degree of (\ref{eqn:mimo_example}) is defined as the vector of minimum integers \(\pmb{r} = \{ r_1 \cdots r_l \}\) such that the following holds w.r.t (\ref{eqn:mimo_example}) $\forall k\in \mathcal{K}, \forall i \in \{ 1 \cdots l \}$:
    $y_i[k+j] = h_i( f(\pmb{x}[k+j-1] ) ), \; \forall j \in \{1 \cdots r_i-1\}$
    $y_i[k+r] =  h_i( f(\pmb{x}[k+r_i-1] ) ) + g(\pmb{x}[k+r_i-1] )\pmb{u}[k+r_i-1]$
\end{definition}

\begin{definition}[Well-Defined Vector Relative Degree]
    Given (\ref{eqn:mimo_example}), we say $\pmb{y}$ has a ``well-defined" VRD w.r.t $\pmb{u}$ if $\exists \; \pmb{u} $ s.t. $\forall \pmb{x}$, $\forall i \in \{ 1 \cdots l \}$, $\frac{\partial y_i [k+r]}{\partial \pmb{u}[k]} \neq 0$.
\end{definition}

If a system does not have a ``well-defined" VRD, then it has a ``poorly defined" VRD. We introduce the notion of $\pmb{\Delta} = \{ \Delta_1 \cdots \Delta_l \} \subset \mathbb{Z}^l$, where $\Delta_i$ is the number of steps backward from $y_i[k+r]$ when the influence of $\pmb{u}[k]$ is lost.

\begin{definition}[$\pmb{\Delta}$]
    If a system has well-defined VRD, then: $\pmb{\Delta} = \{ \Delta_1 \cdots \Delta_l \} = \pmb{r}$. Otherwise, $\forall i \in \{ 1 \cdots l \}, \forall j \in \{ 1 \cdots r-1 \}$, for $\psi_{ij} = \frac{\partial y_i[k+j]}{\partial \pmb{x}[k]}$ and $\mathcal{J}_i = \{ j \in \{ 1 \cdots r_i-1 \} \; | \; \exists \; \pmb{x}[k] \; \text{s.t.} \; \psi_{ij} = 0 \} \cup \{r_i\}$, 
    $\Delta_i = \min j; \text{ s.t } j \in \mathcal{J}_i$.
\end{definition}

\section{Simulation Model}

The system studied in this report is the quadcopter. This is due to it being a non-trivial robotics example, which contains challenging aspects common to many other robotics applications. The states are constrained within bounds, and navigation through space allows for interesting nonlinear constraints such as obstacles.

The state of the quadcopter is $\pmb{x} = \{ x, y, z, q_0, q_1, q_2, q_3, \dot{x}, \dot{y}, \dot{z}, p, q, r, \omega_1, \omega_2, \omega_3, \omega_4 \}$, the input is $\pmb{u} = \{ u_{M1}, u_{M2}, u_{M3}, u_{M4}\}$, and all states and inputs are defined in Table \ref{tab:state_input_description}. Two simulations were developed based on prior work. The first is based on analytically derived equations of motion from \cite{quad_simcon} and the second is built in Mujoco \cite{mujoco} to have a black-box uncertain environment to verify the robustness of the control algorithms implemented \cite{mujoco_quad_sim}.

\begin{center}
    \begin{threeparttable}
        \centering
        \begin{tabular}{l|l}
            \toprule
            \textbf{Symbol} & \textbf{Definition} \\
            \midrule
            \(x, y, z\) & Cartesian positions (m) \\
            \(q_{0,1,2,3}\) & Attitude-defining quaternion (rad) \\
            \(\dot{x}, \dot{y}, \dot{z}\) & Cartesian velocities (m/s) \\
            \(p, q, r\) & Body roll, pitch, yaw rates (rad/s) \\
            \(\omega_{1,2,3,4}\) & Rotor angular velocities (rad/s) \\
            \(\dot{\omega}_{1,2,3,4}\) & Rotor angular accelerations (rad/s/s) \\
            \(u_{M1,2,3,4}\) & Inputs by each motor \\
            \bottomrule
        \end{tabular}
        \caption{List of Quadcopter States and Inputs.}
        \label{tab:state_input_description}
    \end{threeparttable}
\end{center}

In this paper we deem the output of the system to be the quadcopter cartesian coordinates $\pmb{y} = \{x, y, z\}$. In this case, the system has a poorly defined relative degree w.r.t $u_{M1,2,3,4}$. Intuitively this can be understood when considering the quadcopter in a hover state. To move forwards, backward, or left, right, it would first need to rotate before being able to effect acceleration in the desired direction. This is the result of poorly defined VRD. In this paper we will focus on control of the Cartesian positions of this system, largely ignoring orientation control to focus on the poorly defined relative degree in the Cartesian position dynamics.

Throughout this paper, we will use a discretized version of these continuous-time dynamics. Unless otherwise specified an Euler integrator is used such that $\pmb{x}[k+1] = \pmb{x}[k] + \dot{\pmb{x}}[k] T_s$ where $T_s$ is the simulation timestep.

\section{Method}

\begin{figure}[thpb]
    \centering
    \includegraphics[width=0.48\textwidth]{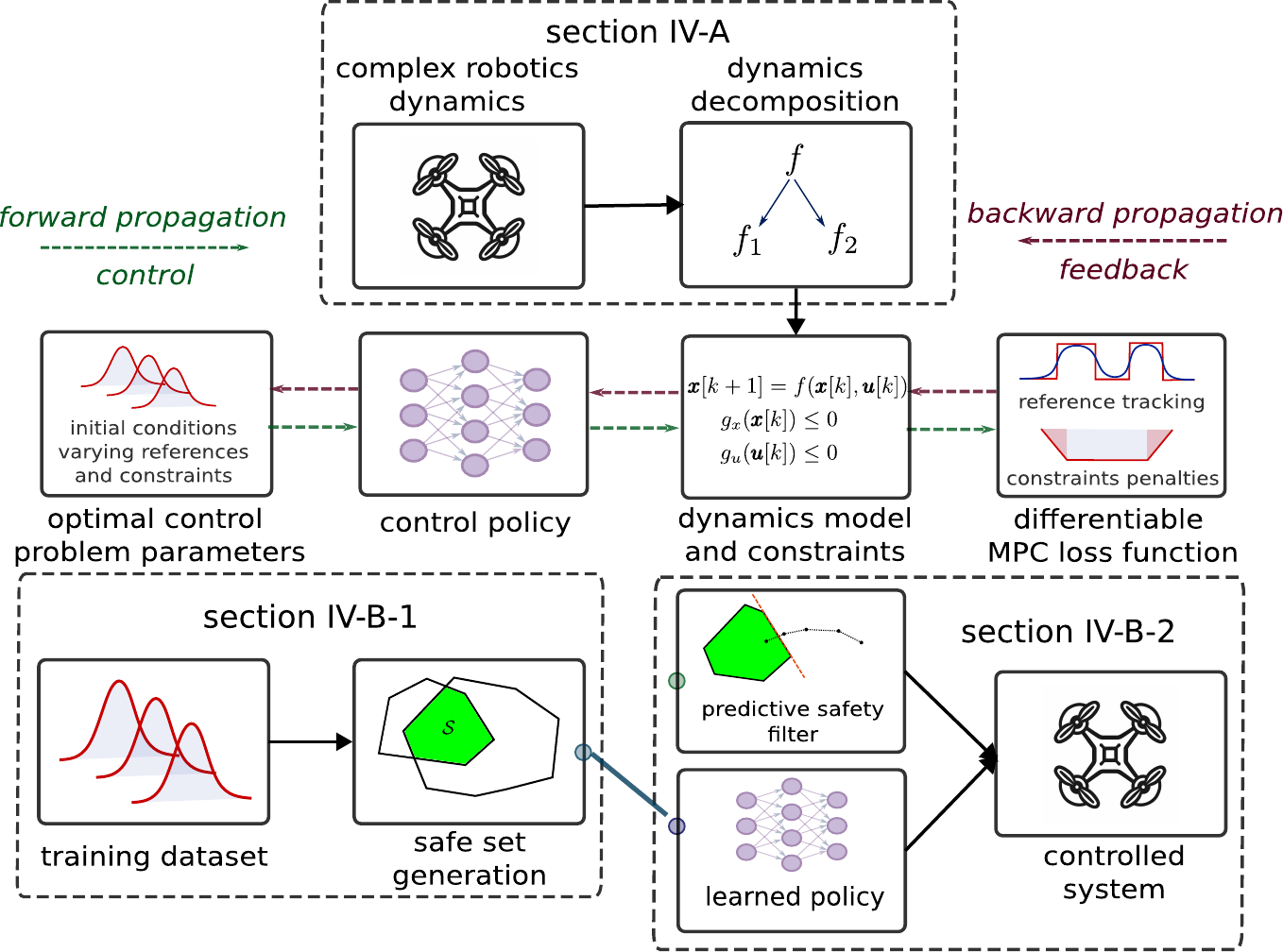}
    \caption{DPC + PSF formulation overview}
    \vspace{-10pt}
    \label{fig:DPC_PSF_overview}
\end{figure}

The method introduced in this paper contains four steps, three of which are novel. We assume we are provided with a system to control with poorly defined VRD, a quadcopter in this case. The first step is to precondition the problem for DPC to train on a system with poorly defined VRD. To do this we introduce a novel systematic way to decompose the dynamics into two subsystems. The first subsystem has well-defined VRD, and the second has either well-defined or poorly-defined VRD. The second step is to train DPC on the first subsystem and save its training dataset. The third step (novel) is to leverage the DPC training dataset to form a data-driven safe set for the DPC closed-loop system. The fourth step (novel) is the only one that takes place online, and it involves running an event-triggered PSF to return the system to its safe set if not within it. An overview of this formulation is found in Figure \ref{fig:DPC_PSF_overview}.

\subsection{Differentiable Predictive Control}

DPC approximates a receding horizon control problem by solving the following parametric optimal control problem (see Table \ref{tab:dpc_symbols} for symbol definitions):

\begin{equation}
    \ell_\text{DPC}(\pmb{x}, \pmb{u}, \pmb{x}_r) = \ell(\pmb{x}, \pmb{u}, \pmb{x}_r) + p_x(\pmb{x}) + p_u(\pmb{u})
    \vspace{-15pt}
    \label{eqn:dpc_cost}
\end{equation}

\begin{equation}
\begin{aligned}
& \underset{\textbf{W}}{\text{min}}
& & \mathbb{E}_{\pmb{x}[0] \thicksim P_{\pmb{x}}, \pmb{x}_r[k] \thicksim P_{\pmb{x}_r}} \, \big( \ell_\text{DPC}(\pmb{x}[k], \pmb{u}[k], \pmb{x}_r[k]) \big) \\ 
& \text{s.t.}
& & \pmb{x}[k+1] = f(\pmb{x}[k], \pmb{u}[k]) \\
& & & \pmb{u}[k] = \pi_\mathbf{w}(\pmb{x}[k], \pmb{x}_r[k]) 
\end{aligned}
\label{eqn:dpc_high_level}
\end{equation}

We obtain the expected value in (\ref{eqn:dpc_high_level}) by sampling.

\begin{center}
    \begin{threeparttable}
        \centering
        \begin{tabular}{l|l}
            \toprule
            \textbf{Symbol} & \textbf{Definition} \\
            \midrule
            \( \mathbb{E} \) & expectation \\
            \( P_{\pmb{x}}, P_{\pmb{r}} \) & training data distributions \\
            \( \ell \) & differentiable cost function \\
            \( (.)^i[k] \) & the $k$th step of the $i$th rollout \\
            \( p_x, p_u \) & state, input constraint penalty functions \\
            \( (f, g) \) & differentiable dynamical system \\
            \(\textbf{W}\) & NN parameters \\
            \( \pi_\textbf{W} \) & NN policy \\
            \( N_r, N_s \) & Number of rollouts and steps respectively \\
            \( \pmb{x}_r \) & reference state \\
            \bottomrule
        \end{tabular}
        \caption{DPC Problem Symbols and Definitions}
        \label{tab:dpc_symbols}
    \end{threeparttable}
\end{center}


A training rollout refers to a time-history of state, input, and reference triplets produced by a simulation in closed-loop using the DPC controller. DPC optimizes $\textbf{W}$ by performing $N_r$ training rollouts of the closed loop system in the forward pass, and evaluating $\ell_\text{DPC}$ across each rollout. Then the NN optimizer uses $\nabla_{\textbf{W}} \ell_{\text{DPC}}$ to update $\pi_\mathbf{W}$ once. This use of $N_r$ rollouts to update the policy once is called a ``batch". We then repeat this process using many ``batches" of size $N_r$ until $\nabla_{\textbf{W}} \ell_{\text{DPC}}$ converges or a maximum number of iterations is reached.

This technique performs best with a well-conditioned $\nabla_{\textbf{W}} \ell_{\text{DPC}}$. This is not the case for dynamics with poorly defined relative degree, such as the quadcopter being investigated in this paper. To fix this issue we developed a novel Algorithm \ref{alg:relative_degree_based_dynamics_decomposition} to decompose a general MIMO nonlinear system into two subsystems, one with guaranteed well-defined VRD, and a second with no guarantees of well-defined VRD. Note that this algorithm can be used to decompose the second subsystem recursively into as many subsystems with well-defined VRD as is required.

\begin{algorithm}
\caption{Relative Degree Dynamics Decomposition}
\label{alg:relative_degree_based_dynamics_decomposition}
\begin{algorithmic}[1] 
    \State \textbf{input}: $\pmb{x}[k+1] = f(\pmb{x}[k]) + g(\pmb{x}[k])\pmb{u}[k]$ \Comment{System}
    \State \textbf{input}: $\pmb{y}[k] = h(\pmb{x}[k])$ \Comment{Output}
    \State $\{ r_1, r_2, ..., r_l \} \gets$ compute number of timesteps $r_i$ for outputs $y_i[k+r_i] \; \forall i \in \{ 1 \cdots l \}$ to be a function of at least one input $\pmb{u}[k]$ 
    \State $\mathcal{I} \gets$ determine set of $i$ for which $\exists \; \pmb{x}[k]$ such that $\frac{\partial y_i[k+r_i]}{\partial \pmb{u}[k]} = 0$ \Comment{Find Poorly Defined Relative Degrees}
    \State $\{ \Delta_1, \Delta_2, \cdots \} \gets$ determine smallest number of steps for influence of $\pmb{u}[k]$ to break down w.r.t output $y_i[k+r_i]$ $\forall i \in \mathcal{I}$
    \State $r_\text{min} \gets \min \Delta_i$ 
    \State $\pmb{x}_{s1} \gets \{ x_a, a \in \{ 1 \cdots n \} \; | \;  \exists \; \pmb{x}, \; j \in \{ 0 \cdots r_\text{min} - 1\}, \; i \in \{ 1 \cdots l \}, \; \text{s.t} \; \frac{\partial y_i[k+j]}{\partial x_a[k]} \neq 0 \}$ \Comment{set of $x$ which appears between $y[k]$ and $y[k+r_\text{min}]$}

    \State $\pmb{u}_{s1} \gets \{ x_a, a \in \{ 1 \cdots n \} \; |  \; x_a \notin \pmb{x}_{s1}, \exists \; \pmb{x}, \; \text{s.t } \frac{\partial y_i[k+r_\text{min}]}{\partial x_a[k]} \neq 0\} \cup \{ u_b, b\in \{ 1 \cdots m \} \; | \; \exists \; \pmb{x} \; \text{s.t } \frac{\partial y_i[k+r_\text{min}]}{\partial u_b[k]} \neq 0  \}$    
    \State $\pmb{x}_{s2} \gets \{ x_a, a \in \{ 1 \cdots n \} \; | \; x_a \notin \{ \pmb{x}_{s1} \cup \pmb{u}_{s1} \} \}$ 
    \State $\pmb{u}_{s2} \gets \{ u_b, b\in \{1\cdots m \} \; | \; u_b \notin \pmb{u}_{s1} \}$
    \State $\{ f_{s1}, g_{s1}, f_{s2}, g_{s2} \} \gets$ determine equations governing $\{ \pmb{x}_{s1}, \pmb{x}_{s2} \}$ from $f, g$
    \State $\pmb{x}_\text{s1}[k+1] = f_\text{s1}(\pmb{x}_\text{s1}[k]) + g_\text{s1}(\pmb{x}_\text{s1}[k])\pmb{u}_\text{s1}[k]$ \Comment{Form subsystem 1}
    \State $\pmb{x}_\text{s2}[k+1] = f_\text{s2}(\pmb{x}_\text{s2}[k]) + g_\text{s2}(\pmb{x}_\text{s2}[k])\pmb{u}_\text{s2}[k]$ \Comment{Form subsystem 2}
\end{algorithmic}
\end{algorithm}

Applying this algorithm to the quadcopter dynamics yields two subsystems $(f_{s1}, g_{s1})$ and $(f_{s2}, g_{s2})$. The first subsystem uses state $\pmb{x}_{s1} = \{ x, y, z, \dot{x}, \dot{y}, \dot{z}\}$ with input $\pmb{u}_{s1} = \{ \ddot{x}, \ddot{y}, \ddot{z} \}$. The second subsystem uses state $\pmb{x}_{s2} = \{ q_0, q_1, q_2, q_3, p, q, r, \omega_{M1}, \omega_{M2}, \omega_{M3}, \omega_{M4}\}$ and input $\pmb{u}_{s2} = \{ u_1, u_2, u_3, u_4 \}$. 

We use DPC to control subsystem 1, and a cascade Proportional (P) controller for subsystem 2. This method of splitting the dynamics and utilizing a computationally cheap classical control method for controlling subsystem 2 is similar to how many remote-controlled quadcopters work today. This practical control architecture combining DPC and P controllers, lends itself well to field tests in future work.

\subsection{DPC with Predictive Safety Filter}

In this section, we introduce the Predictive Safety Filter (PSF) that utilizes a data-driven safe set produced by the DPC training procedure. The PSF is then applied to the DPC control to enforce safety.

\subsubsection{Safe Set Generation}
During the offline DPC policy optimization, large numbers of simulations are sampled using suboptimal policies. This is a significant dataset representing regions of the state space that are known to the policy. Sampling-based performance verification techniques grow in efficacy with the size of the dataset they are provided. In Algorithm~\ref{alg:safe_set_generation}, we show a technique to utilize the DPC training dataset to generate a data-driven safe set for the closed loop system. All relevant symbols are defined in Table \ref{tab:safe_set_symbols}.

\begin{algorithm}
\caption{Safe Set Generation}
\label{alg:safe_set_generation}
\begin{algorithmic}[1] 
    \State \textbf{input}: DPC training rollouts
    \State \textbf{input}: Non-convex constraints $\mathbb{V}_i \; \forall i \in \{ 1 \cdots N_v \}$
    \State $\mathbb{L} \gets$ \{ rollout $\in$ training rollouts $|$ rollout satisfies constraints and converges to reference \}
    \Function{Conv}{dataset}
        \State \textbf{return} convex hull of points in dataset
    \EndFunction
    \Function{FindTransform}{$\mathbb{V}_i$}
        \State $\mathbb{T}_i \gets$ Find transform to convexify $\mathbb{V}_i$
        \State \textbf{return} $\mathbb{T}_i$ 
    \EndFunction
    \State $\mathcal{S}_\text{cvx} \gets \text{Conv}(\mathbb{L})$
    \For{$i \in \{ 1 \cdots N_v\}$}
        \State $\mathbb{T}_i \gets \text{FindTransform}(\mathbb{V}_i)$
        \State $\mathbb{L}_{\mathbb{V} i} \gets \mathbb{T}_i(\mathbb{L})$
        \State $\mathcal{S}_{\mathbb{V} i} \gets \text{Conv}(\mathbb{L}_{\mathbb{V} i})$
    \EndFor
    \State $\mathcal{S} \gets \mathcal{S}_\text{cvx} \cap \left(\bigcap_{i=1}^{N_v} \mathcal{S}_{\mathbb{V}i}\right)$
    
\end{algorithmic}
\end{algorithm}

\begin{center}
    \begin{threeparttable}
        \centering
        \begin{tabular}{l|l}
            \toprule
            \textbf{Symbol} & \textbf{Definition} \\
            \midrule
            \( \mathbb{L} \) & All training rollouts s.t. constraints satisfied \\ 
            \( \mathbb{V}_i \) & $i$th non-convex constraint \\
            \( \mathbb{L}_{\mathbb{V} i} \) & \( \mathbb{L} \) transformed s.t. $\mathbb{V}_i$ is convex\\
            \( \mathcal{S} \) & safe set \\
            \( \mathcal{S}_\text{cvx} \) & safe set for all convex constraints \\
            \( \mathcal{S}_{\mathbb{V} i} \) & safe set for the $i$th non-convex constraint \\
            \bottomrule
        \end{tabular}
        \caption{Safe Set Symbols and Definitions}
        \label{tab:safe_set_symbols}
    \end{threeparttable}
\end{center}

This algorithm makes two primary assumptions. The first is that if a given iteration of the policy satisfies the constraints/convergence criteria mentioned above across a given rollout, then a subsequent iteration of the policy with lower $\ell_{\text{DPC}}$ will also satisfy the criteria from any point along the rollout. The second assumption is that, in the absence of non-convex constraints, any point in a convex hull of safe points is indeed safe, which we can only guarantee with infinite training rollouts to cover every possible point in the state space.

Note, the denser the points, the more accurate the second assumption is. This is not rigorous, and indeed finding and validating safety certificates across continuous state spaces in $\mathbb{R}^n$ that do not require the second assumption remains an open research area \cite{neural_lyapunov_control}. However, there is precedent to use sampling-based methods for similar problems \cite{hj_reachability}.

To better understand the algorithm let us introduce a non-convex constraint to $f_{s1}$, a cylindrical obstacle of given radius $r$, positioned at $(x_\text{cyl}, y_\text{cyl})$ with infinite $z$ height. This constraint takes the form $r^2 \leq (x-x_\text{cyl})^2 + (y-y_\text{cyl})^2$. Firstly we process $\mathbb{L}$ to filter out trajectories that violate any constraints, in this case just the cylinder constraint and the quadcopter box state and input constraints. We generate $\mathcal{S}_\text{cvx}$ by using the QuickHull \cite{quick_hull} algorithm applied to $\mathbb{L}$. Note this is tractable due to the low dimensionality of subsystem 1. If subsystem 1 had a much higher dimensionality, there exist algorithms that have complexity $\mathcal{O}(N^2)$ w.r.t dimensionality $N$ \cite{high_dimension_cvx_hull} that could be used. Then we must find the transform to convexify the single non-convex constraint defined by the cylinder obstacle. This is simple in this case through the use of cylindrical coordinates. We convert $\pmb{x}_{s1} = \{ x, y, z, \dot{x}, \dot{y}, \dot{z} \}$ into $\pmb{x}_c = \{ x_c, \dot{x}_c \}$, where $x_c$, $\dot{x}_c$ represent distance from and velocity away from the cylinder respectively. The transform $\mathbb{T}_1$ is shown in (\ref{eqn:cyl_constraint_state_transform}), where $\pmb{x}_c = \mathbb{T}_1(\pmb{x}_{s1})$. In this transformed space the cylinder constraint takes the convex form: $x_c > 0$.

\begin{equation}
    \begin{aligned}
        x_{c} &= \sqrt{(x - x_\text{cyl})^2 + (y - y_\text{cyl})^2} - r \\
        \dot{x}_{c} &= \dot{x}\frac{x - x_\text{cyl}}{x_c + r} + \dot{y}\frac{y - y_\text{cyl}}{x_c + r}
    \end{aligned}
    \label{eqn:cyl_constraint_state_transform}
\end{equation}

$\mathcal{S}_{v1}$ is then formed by applying QuickHull to the transformed dataset. Finally $\mathcal{S}$ is formed by forming the intersection of $\mathcal{S}_\text{cvx}$ and $\mathcal{S}_{v1}$. Note that although this forms a non-convex set in the original untransformed space, each safe set that we use to form the intersection can be queried independently in their transformed space to determine if a point lies within $\mathcal{S}$. This is the key to allowing us to leverage efficient convex-hull-based algorithms for a non-convex problem.

\begin{figure}[thpb]
    \centering
    \includegraphics[width=0.48\textwidth]{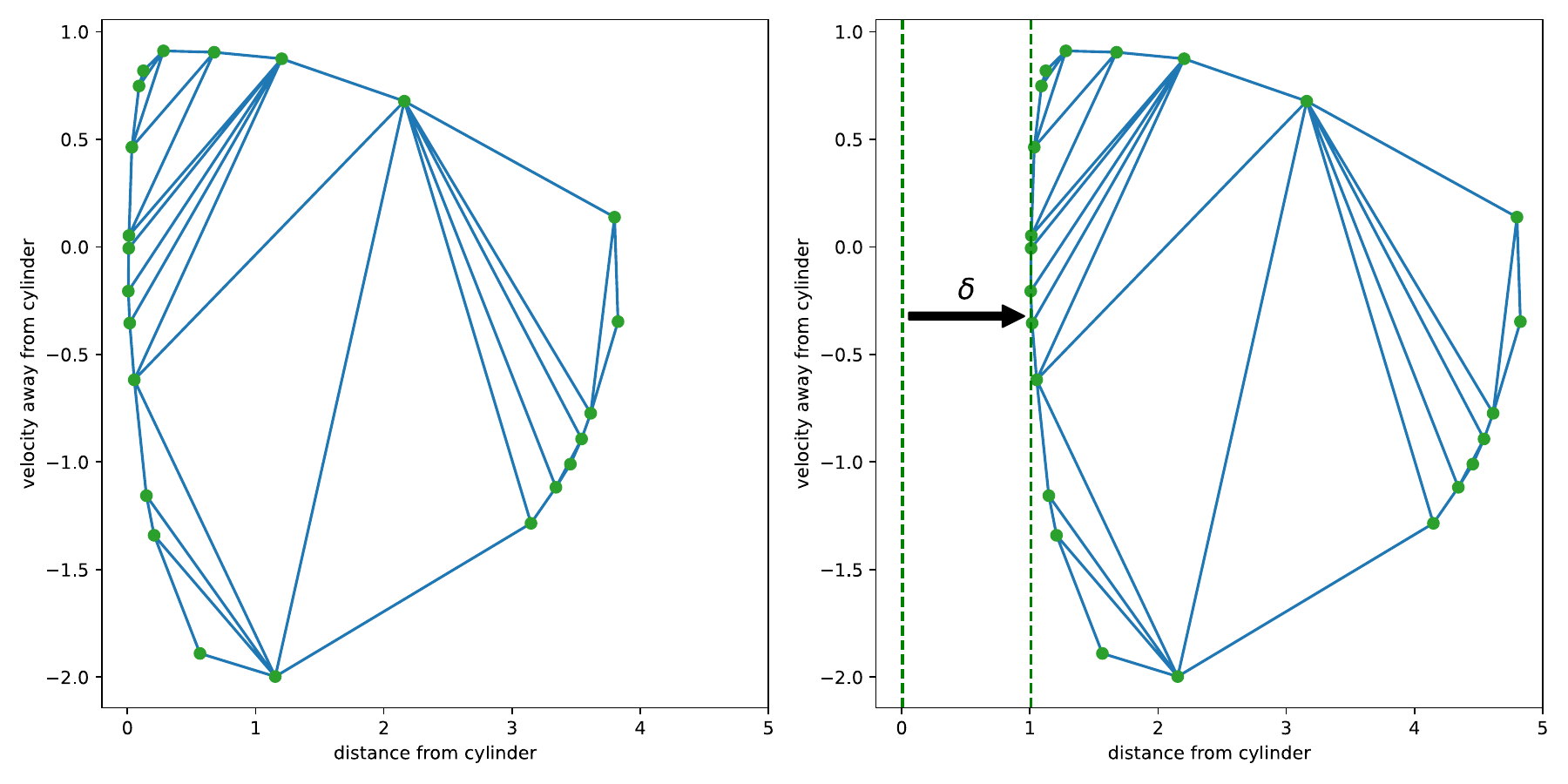}
    \caption{Robust $\delta$ Applied to $\mathcal{S}_{\mathbb{V}1}$}
    \vspace{-10pt}
    \label{fig:f_s1_cvx_hull_shift}
\end{figure}

Robustness margins are known to be effective for the application of predictive control techniques with plant model mismatch \cite{robust_psf_wences}. In this case, because the DPC-controlled subsystem 1 is unaware of the closed-loop dynamics of subsystem 2 a robustness term $\delta$ is added to $\mathcal{S}$ which shifts the vertices of $\mathcal{S}_{\mathbb{V}1}$ to make it a more conservative estimate w.r.t the cylinder constraint as shown in Figure \ref{fig:f_s1_cvx_hull_shift}. This mismatch created by DPC not being aware of subsystem 2 can also cause a constant offset when tracking a time-varying reference due to the delay subsystem 2 adds to subsystem 1. Other robustness margins have shown promise in preliminary testing, such as creating a new convex hull by shifting the vertices of the original towards the centroid of the hull. However, a simple shift suffices for the single affine inequality constraint in $f_{s1}$.

\subsubsection{Application of Safe Set}

The Predictive Safety Filter (PSF) was implemented to approximately enforce forward invariance towards the safe set. To plan to minimize the difference between a filtered control action and an unfiltered action, a representation of the control law is required. Directly using the NN representation of the policy creates an optimization landscape that is difficult for line search-based methods (such as IPOPT \cite{ipopt}) to traverse. Therefore a first-order approximation to the policy is used at every simulation timestep as shown in (\ref{eqn:nn_taylor}), where $\pmb{u}_{s1}^\text{jac}(\pmb{x}_{s1}[k])|_{\pmb{x}_{s1}[0]}$ represents the first order approximation of the policy around $\pmb{x}_{s1}[0]$ evaluated at $\pmb{x}_{s1}[k]$, and $\mathbb{J}( \pmb{u}_{s1}^{\text{dpc}})|_{\pmb{x}_{s1}[0]}$ represents the Jacobian of the policy around $\pmb{x}_{s1}[0]$.

\begin{equation}
\begin{aligned}
    \pmb{u}_{s1}^\text{jac}(\pmb{x}_{s1}[k])|_{\pmb{x}_{s1}[0]} &= \pmb{u}_{s1}^{\text{dpc}}(\pmb{x}_{s1}[0]) \\
    & + \mathbb{J}( \pmb{u}_{s1}^{\text{dpc}})|_{\pmb{x}_{s1}[0]} \cdot (\pmb{x}_{s1}[k] - \pmb{x}_{s1}[0])
\end{aligned}
\label{eqn:nn_taylor}
\end{equation}

We also want the PSF to converge the system towards the safe set. The naive application of every halfspace that defines the intersection of convex hulls defining the safe set is computationally intractable for complex geometries. Therefore we only consider the closest face of each convex hull at every timestep. Intuitively this is a linearization of the convex hull constraint - an approximation of the tangent plane of the convex hull at the projection of $\pmb{x}_{s1}[0]$ onto its surface. Finding the nearest face naively by searching across a set of precomputed faces is memory inefficient in high dimensions, as the number of faces is much larger than the number of vertices that define them. Therefore the approach taken here is to find the nearest $n$ non-colinear points for an $n$ dimensional convex hull, and consider the hyperplane that passes through all of them as the nearest face of the hull. We then ensure that the normal of this hyperplane is pointed towards the centroid of its convex hull. This process is done across all convex hulls that define the safe set, and then for each nearest hyperplane we minimize the term $\pmb{x}_{s1}[k]^T\pmb{w}_i + b_i$, where $\pmb{w}_i$ and $b_i$ are the normal and offset defining each hyperplane respectively. This makes the minimization of the term approximately represent movement towards the center of the convex hull. However, it is only desired that the PSF brings the system within the hull by some margin $m_i$, and so a Softplus function is applied ($\text{Softplus}(x) = \log(1+e^x)$) to reduce the weight of the term once we are within the hull. The Softplus function was chosen over alternatives due to its continuous differentiability, and this results in the final formulation shown in (\ref{eqn:psf_formulation}).

\begin{equation}
    \begin{aligned}
        \mathcal{P}_\text{sf}(\pmb{x}_{s1,0}, \pmb{w}_{1\cdots N_v}, &b_{1\cdots N_v}) = \min_{\pmb{x}_{s1},\pmb{u}_{s1}} \sum_{k=0}^{N} \sum_{i=1}^{N_v} \\ 
        & \left\| \pmb{u}_{s1}^\text{jac}(\pmb{x}_{s1}[k])|_{\pmb{x}_{s1}[0]} - \pmb{u}_{s1}[k] \right\|_2 \\
        & + \alpha_i \; \text{Softplus}(\pmb{x}_{s1}[k]^T \pmb{w}_i + b_i + m_i) \\
        \text{s.t.:} \quad \pmb{x}_{s1}[k+1] &= f_{s1}(\pmb{x}_{s1}[k]) + g_{s1}(\pmb{x}_{s1}[k])\pmb{u}_{s1}[k], \\
        \quad \pmb{x}_{s1}[0] &= \pmb{x}_{s1,0}.
    \end{aligned}
    \label{eqn:psf_formulation}
\end{equation}

\begin{center}
    \begin{threeparttable}
        \centering
        \begin{tabular}{l|l}
            \toprule
            \textbf{Symbol} & \textbf{Definition} \\
            \midrule
            \( \pmb{u}_{s1}^\text{dpc} \) & nominal DPC control being filtered \\
            \( \pmb{u}_{s1}^\text{jac} \) & first order approximation of $\pmb{u}_{s1}^\text{dpc}$ \\
            \( \pmb{u}_{s1}^\text{sf} \) & safety filtered $\pmb{u}_{s1}^\text{fo}$ \\
            \( \pmb{w}_i, b_{i} \) & hyperplane normal and constant offset \\
            \( m_i \) & convex hull minimization margin term \\ 
            \( \alpha_i \) & convex hull weighting term \\
            \bottomrule
        \end{tabular}
        \caption{Safety Filter Symbols and Definitions}
        \label{tab:sf_symbols}
    \end{threeparttable}
\end{center}

Another technique used for accelerating computation was an event trigger for the activation of this PSF. If we are within the safe set the nominal $\pmb{u}_{s1}^\text{dpc}$ produced by the DPC is used directly without filtering.

\begin{algorithm}
\caption{Event-Triggered PSF}
\label{alg:triggered_safety_filter}
\begin{algorithmic}[1] 
    \State \textbf{input}: $\pmb{x}_{s1,0}$, $\pmb{u}_{s1}^\text{dpc}$, $\mathcal{S}$
    \Function{d}{$\pmb{p}_1, \pmb{p}_2$}
        \State \textbf{return} $||\pmb{p}_1 - \pmb{p}_2||_2$ \Comment{Euclidean Distance}
    \EndFunction
    \Function{$\mathbb{H}$}{$\pmb{x}, \mathcal{S}$}
        \State $\{ \pmb{p}_1 \cdots \pmb{p}_{Ns}\} \gets$ sort $\pmb{p}_i \in \mathcal{S}$ by D($\pmb{x}, \pmb{p}_i$) \Comment{sort by Euclidean distances, D$(\pmb{x},\pmb{p}_1)$ $<$ D($\pmb{x},\pmb{p}_2$) $<$ $\cdots$ $<$ D($\pmb{x},\pmb{p}_{Ns}$)}
        \State $\mathcal{P} \gets \{ \pmb{p}_i | \nexists \text{ colinear } (\pmb{p}_i, \pmb{p}_j) \; \forall \; (i,j) \in \{ 1 \cdots n\} \} \subseteq \{ \pmb{p}_1 \cdots \pmb{p}_{Ns}\}$ \Comment{determine first $n$ non colinear points}
        \State $\pmb{w}$, $b$ $\gets$ hyperplane defined by $\mathcal{P}$ \Comment{ensure that $\pmb{w}$ is pointed towards the centroid of $\mathcal{S}$ and is normalized}
        \State \textbf{return} $\pmb{w}$, $b$
    \EndFunction
    \State $\pmb{w}_1$, $b_1$ $\gets \mathbb{H}(\pmb{x}_{s1,0}, \mathcal{S_\text{cvx}})$
    \For{$i \in \{ 1 \cdots N_v \}$}
        \State $\pmb{w}_{i+1}, b_{i+1} \gets \mathbb{H}(\mathbb{T}_i(\pmb{x}_{s1,0}), \mathcal{S}_{\mathbb{V}_i})$
    \EndFor
    \If{$\pmb{w}_1^T \pmb{x}_{s1,0} + b_1 \leq 0 \And \pmb{w}_{i+1}^T \mathbb{T}_i(\pmb{x}_{s1,0}) + b_{i+1} \leq 0 \; \forall i \in \{ 1 \cdots N_v+1 \}$}
    \State $\pmb{u}_{s1}^\text{sf} \gets \pmb{u}_{s1}^\text{dpc}$ \Comment{$\pmb{x}_{s1,0} \in \mathcal{S}$}
    \Else
        \State $\pmb{u}_{s1}^\text{sf} \gets \mathcal{P}_\text{sf}(\pmb{x}_{s1,0}, \pmb{w}_1, \pmb{w}_2, b_1, b_2)$
    \EndIf
\end{algorithmic}
\end{algorithm}

\begin{figure}[thpb]
    \centering
    \includegraphics[width=0.48\textwidth]{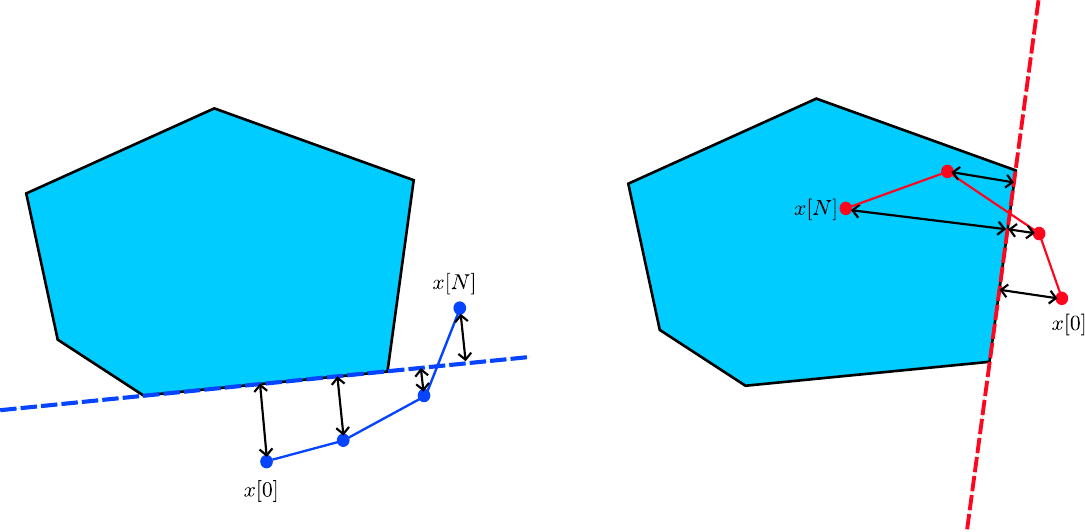}
    \caption{Illustration of a single face of a safe set convex hull being used as a constraint in the PSF at every timestep}
    \vspace{-8pt}
    \label{fig:safe_set_predictive_constraints}
\end{figure}

The PSF becomes more effective the longer its prediction horizon is, but this comes at the cost of computation time. Predictions far into the future are of less importance than those closer in time. Therefore it was decided to use a linearly increasing timestep prediction horizon due to its simple closed form solution of $\delta t[k] = T_s + 2k \frac{\frac{T_f}{N} - T_s}{N - 1}$, where $T_s$ is the simulation timestep, $T_f$ is the final time to be reached by the horizon, $N$ is the number of steps allowed, and $k$ is the timestep along the prediction horizon. This linearly increasing prediction timestep loses accuracy with its length, but with $N = \frac{T_f}{T_s}$, $\delta t[k] = T_s$ a standard constant timestep prediction horizon is recovered, and so a large enough $N$ must be chosen to ensure the accuracy is enough for our purposes.

In summary, the full DPC + PSF formulation has been described, with the first step being the use of Algorithm \ref{alg:relative_degree_based_dynamics_decomposition} to decompose dynamics into a well-defined problem for DPC to control. The second step is using the DPC training dataset to form the safe set as shown in Algorithm \ref{alg:safe_set_generation}, and the third step is to integrate the DPC controller and safe set into the event-triggered PSF formulation shown in Algorithm \ref{alg:triggered_safety_filter}. The next section presents experimental results and compares the final controller with MPC.

\section{Experiments}

We will compare the performance of DPC and DPC + PSF with that of two MPC controllers. The DPC controller is trained using $\ell_\text{MPC}$ in (\ref{eqn:mpc_cost}). The first MPC controller utilizes the same variable timestep prediction horizon technique as the DPC + PSF, and will be referred to as Variable Timestep Nonlinear MPC (VTNMPC). The other MPC controller is simply Nonlinear MPC (NMPC). All controllers utilize warm starting of their optimization variables if they have any. Three scenarios were chosen for comparison to represent important applications for quadcopter navigation. The first task is navigation around an obstacle, the second task is tracking a reference trajectory, and the third task is navigation around an obstacle with challenging initial conditions. All controllers utilize the same internal model, and all controllers are applied to the uncertain Mujoco environment to incorporate plant-model mismatch. The metrics used to evaluate the efficacy of each controller on any task are the MPC cost $\ell_\text{MPC}$ in (\ref{eqn:mpc_cost}), and the total closed loop simulation computation time. The MPC cost used $Q=\text{diag}[1,1,1,0,0,0,0,1,1,1,1,1,1,0,0,0,0]$ and $R=\text{diag}[1,1,1,1]$, this means that zero weighting is given to the orientation of the quadcopter and the angular velocity of the rotors, all other states and inputs are equally weighted. This was chosen because quadcopter orientation is unimportant to navigation, and quaternion error adds unecessary complexity. All simulations are conducted with a timestep of 1ms and are controlled at every timestep. All tests are run on Ubuntu 22.04 using a Ryzen 5800X3D.

\begin{equation}
    \ell_\text{MPC}(\pmb{x}, \pmb{u}, \pmb{x}_r) = (\pmb{x}_r-\pmb{x})^T Q (\pmb{x}_r-\pmb{x}) + \pmb{u}^T R \pmb{u}
    \label{eqn:mpc_cost}
\end{equation}

\subsection{Navigation}

In this task, the controllers are tasked with navigating to a target position in the presence of a cylindrical obstacle. The rollouts were created over a simulation period of 5.0s, preventing the quadcopter from remaining at the desired location for a long period. Both MPC controllers required additional tuning to achieve satisfactory results due to the plant-model mismatch. It was found that the cylinder constraint had to be increased in size along the prediction horizon to add robustness. No such modification was required for the DPC or DPC + PSF formulation. The augmented cylinder constraint was: $r^2 m \leq (x_\text{cyl} - x)^2 + (y_\text{cyl} - y)^2$, where $m = 1 + 0.1 t$, where $t$ is the time delta between the start of the prediction horizon and the current timestep in the prediction horizon, $x_\text{cyl}, y_\text{cyl}$ are the $(x,y)$ coordinates of the center of the cylinder, and $r$ is the radius of the cylinder.

A prediction horizon of at least 2.0s was found to be required for reliable avoidance of the cylinder obstacle, otherwise, the MPC could find it difficult to plan a path around the cylinder. Therefore all controllers used prediction horizons of 2.0s, the NMPC subsequently used $N=2000$, and both the DPC + PSF and VTNMPC used an arbitrarily chosen $N=30$.

The PSF only engaged briefly during this task, meaning that the DPC-controlled system almost never left its safe set, implying that the DPC training dataset was representative of this task. An unexpected result is that the DPC + PSF outperformed the VTNMPC controller in terms of true MPC cost on average. We hypothesize that this is because of the substantial plant-model mismatch of the VTNMPC due to its inaccurate long-term predictions, which result in large control actions and contribute to the large cost. 

\subsection{Reference Trajectory Tracking}

\begin{figure}[thpb]
    \centering
    \includegraphics[width=0.5\textwidth]{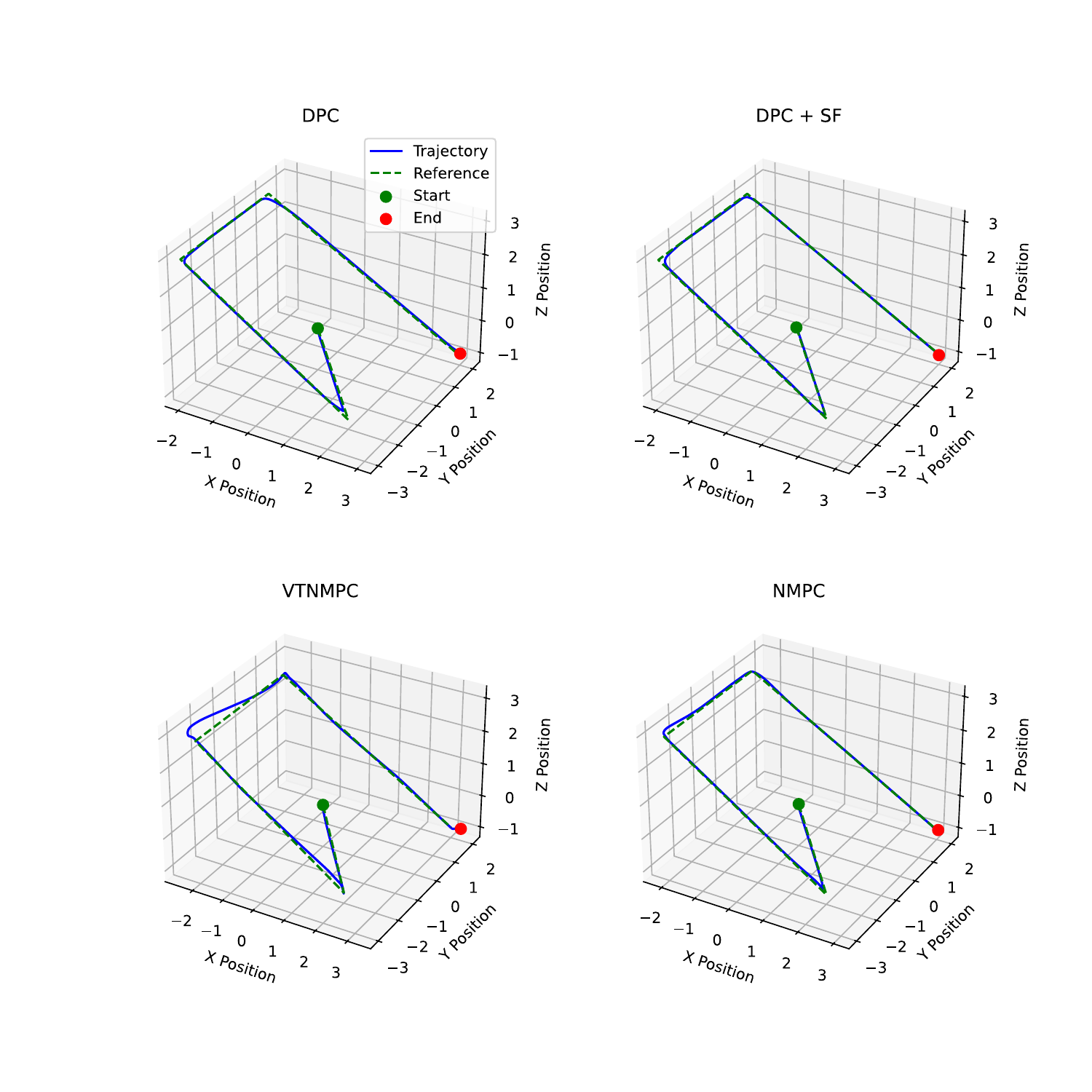}
    \caption{Trajectory Tracking Task}
    \vspace{-15pt}
    \label{fig:traj}
\end{figure}

In this task, a time-varying reference trajectory is generated by linearly interpolating between waypoints over time, and the controllers are tasked with tracking it. This task is longer at 20s and contains only convex constraints and therefore was found to not require a long prediction horizon for good MPC results. We chose a prediction horizon of 0.5s for all controllers. This makes the NMPC $N=500$, and the DPC + PSF and VTNMPC $N$ were kept the same at $N=30$. In theory this should make NMPC much more competitive, which is a result we see. Figure \ref{fig:traj} appears to demonstrate DPC and DPC + PSF having superior performance to NMPC and VTNMPC, but the MPC cost in Table \ref{tab:performance_comparison} does not seem to reflect this. This is because DPC and DPC + PSF controlled quadcopters have a temporal offset ahead and behind the time-varying reference trajectory throughout the simulation. This is due to the low-level P controlled dynamics that the DPC does not consider. This has already been compensated for by offsetting the reference provided to the DPC-based controllers in time, as the full reference is known a-priori, but without further fine-tuning of the DPC-based controllers with knowledge of the low-level P controller, this was the best MPC cost achieved.

\begin{figure}[thpb]
    \centering
    \includegraphics[width=0.45\textwidth]{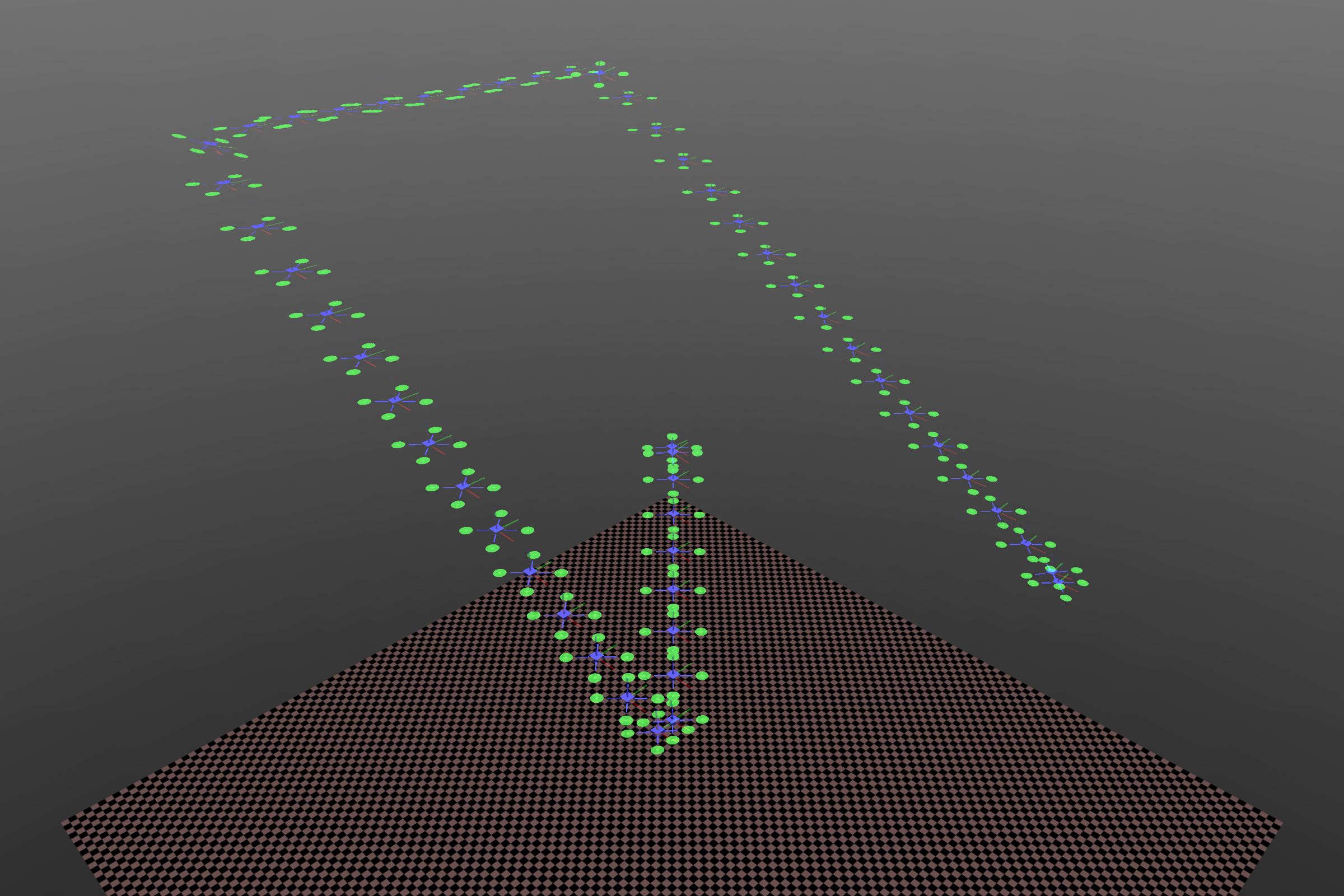}
    \caption{Trajectory Tracking Task Mujoco Render}
    \vspace{-10pt}
    \label{fig:nav}
\end{figure}

\subsection{Adversarial Navigation}

\begin{figure}[thpb]
    \centering
    \includegraphics[width=0.5\textwidth]{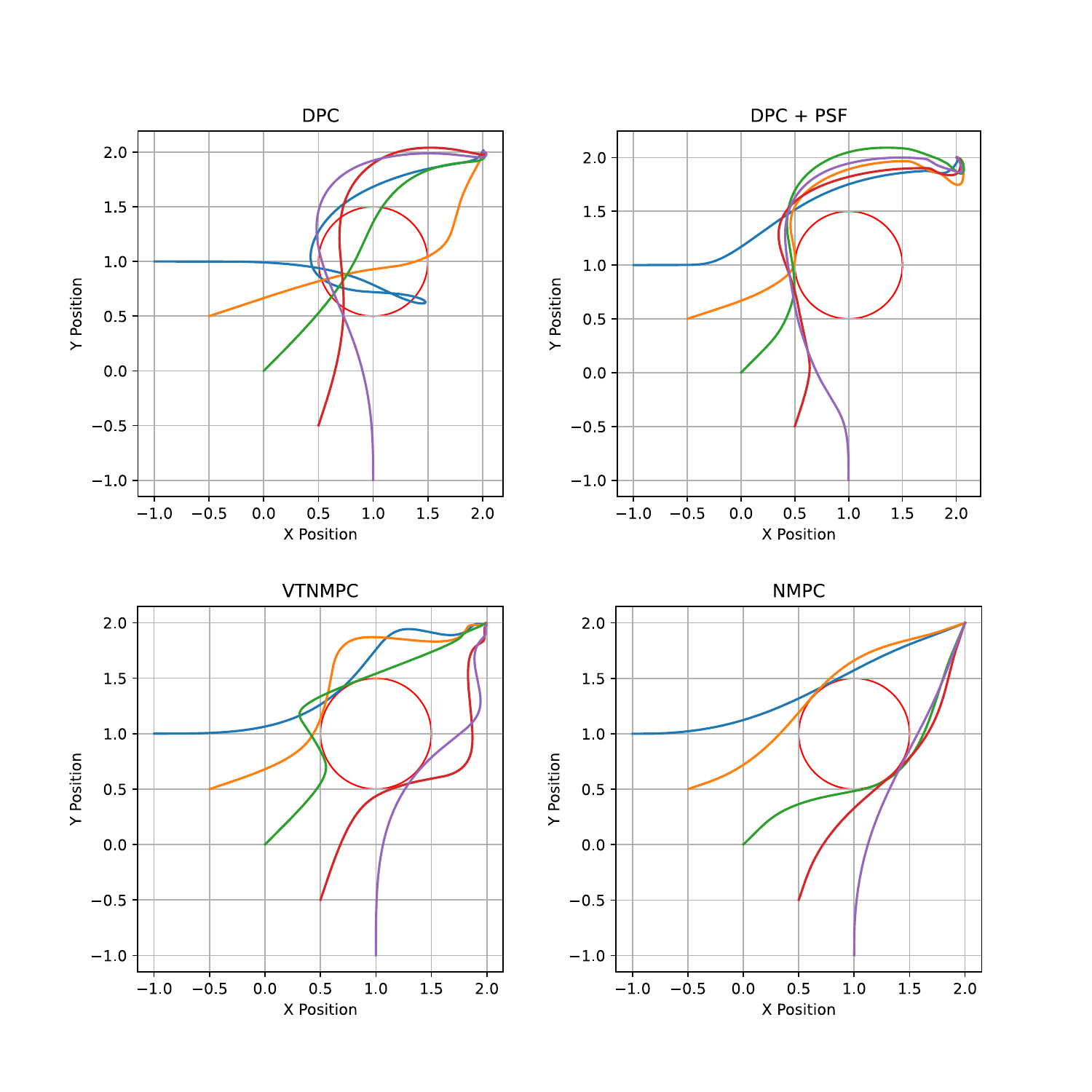}
    \caption{Adversarial Navigation Task}
    \vspace{-15pt}
    \label{fig:adv_nav}
\end{figure}

In the navigation and trajectory tracking tasks, the DPC-based controllers were trained with a dataset that was representative of the task, therefore the safe set and PSF were used very sparingly. In this task, we initialize the quadcopter at the same initial positions as in the navigation task, but instead of starting from hover, the quadcopter is given an initial velocity of 2.25 m/s directly towards the center of the cylinder. This initial condition is significantly outside of the DPC training dataset and therefore requires the PSF to ensure constraint satisfaction. This can be seen in Figure \ref{fig:adv_nav}, where the quadcopter does not avoid the cylinder successfully, resulting in an infinite $\ell_\text{MPC}$. However here we see the first demonstration that DPC + PSF has successfully converged the quadcopter back into a safe set for the DPC controller by its successful avoidance of the cylinder.

\subsection{Results}

The final result shows that in all three scenarios, the DPC + PSF formulation satisfied all constraints, took an order of magnitude less computation time than VTNMPC, and three orders of magnitude less than NMPC. This was whilst having a \textbf{lower} MPC cost than VTNMPC in the navigation task, and having an MPC cost less than an order of magnitude larger than VTNMPC and NMPC \textbf{in all tasks}.

\begin{table*}[t]
\begin{center}
    \begin{threeparttable}
        \centering
        \begin{tabular}{ll|llll}
            \toprule
            Task & Criterion & DPC & DPC + PSF & VTNMPC & NMPC \\
            \midrule
            \multirow{2}{*}{\textbf{Navigation (5s)}} & MPC Cost & 18677.00 & 19716.85 & 27441.92 & 16202.96 \\
            & Simulation Time (s) & \pmb{11.34} & 31.70 & 329.58 & 40696.99 \\
            \midrule
            \multirow{2}{*}{\textbf{Trajectory Tracking (20s)}} & MPC Cost & 20164.08 & 20085.40 & 6972.52 & 2046.60 \\
            & Simulation Time (s) & \pmb{47.51} &  94.52 & 1070.92 & 33298.85 \\
            \midrule
            \multirow{2}{*}{\textbf{Adversarial Navigation (10s)}} & MPC Cost & $\infty$ & 27556.22 & 13323.20 & 11134.36 \\
            & Simulation Time (s) & 23.03 & \pmb{72.93} & 570.58 & 60649.51 \\
            \bottomrule
        \end{tabular}
        \caption{Experimental Performance Comparison}
        \label{tab:performance_comparison}
    \end{threeparttable}
\end{center}
\end{table*}

\section{Conclusion}

In this paper, we have presented a novel technique for promoting the safety of a DPC-controlled robotics system in a data-driven manner, at three orders of magnitude less computational time than state-of-the-art MPC systems and having comparable performance as measured by the MPC cost function. Safety is not guaranteed due to only considering the approximate tangent plane of the safe sets in the PSF formulation. Achieving guaranteed safety is an area of further research. The computational cost of our technique is approaching real-time realizability on highly mobile robotics, something that MPC formulations have struggled to do. In the future, we will explore improving performance in multi-modal situations, extending DPC to the full system dynamics, and applications to other robotics platforms.

\addtolength{\textheight}{-1cm}   


\bibliographystyle{IEEEtran}
\bibliography{lib}

\end{document}